\documentclass[10pt,twocolumn,letterpaper]{article}

\usepackage{cvpr}              % To produce the CAMERA-READY version
\usepackage{graphicx}
\usepackage{amsmath}
\usepackage{amssymb}
\usepackage{booktabs}
\usepackage{caption}
\usepackage{relsize}
\usepackage[pagebackref,breaklinks,colorlinks]{hyperref}
\usepackage[accsupp]{axessibility}

\usepackage[capitalize]{cleveref}
\crefname{section}{Sec.}{Secs.}
\Crefname{section}{Section}{Sections}
\Crefname{table}{Table}{Tables}
\crefname{table}{Tab.}{Tabs.}

\def\cvprPaperID{3} % *** Enter the CVPR Paper ID here
\def\confName{VAND}
\def\confYear{2023}

\begin{document}

\title{Multi-Task Learning based Video Anomaly Detection with Attention}

\author{Mohammad Baradaran\\
Université Laval, Canada\\
{\tt\small mohammad.baradaran.1@ulaval.ca}
\and
Robert Bergevin\\
Université Laval, Canada\\
{\tt\small robert.bergevin@gel.ulaval.ca}
}
\maketitle

%%%%%%%%% ABSTRACT
\begin{abstract}
   Multi-task learning based video anomaly detection methods combine multiple proxy tasks in different branches to detect video anomalies in different situations. Most existing methods suffer from one of these shortcomings: I) Combination of proxy tasks in their methods is not in a complementary and explainable way. II) Class of the object is not effectively considered. III) All motion anomaly cases are not covered. IV) Context information is not engaged in anomaly detection. To address these shortcomings, we propose a novel multi-task learning based method that combines complementary proxy tasks to better consider the motion and appearance features. In one branch, motivated by the abilities of the semantic segmentation and future frame prediction tasks, we combine them into a novel task of future semantic segmentation prediction to learn normal object classes and consistent motion patterns, and to detect respective anomalies simultaneously. In the second branch, we leverage optical flow magnitude estimation for motion anomaly detection and we propose an attention mechanism to engage context information in normal motion modeling and to detect motion anomalies with attention to object parts, the direction of motion, and the distance of the objects from the camera. Our qualitative results show that the proposed method considers the object class effectively and learns motion with attention to the aforementioned determinant factors which results in precise motion modeling and better motion anomaly detection. Additionally, quantitative results show the superiority of our method compared with state-of-the-art methods.
\end{abstract}

%%%%%%%%% BODY TEXT
\section{Introduction}
\label{sec:intro}

With the growth of surveillance cameras, automatic analysis of video content is called for. Generally, the aim of this analysis is to detect anomalous events (\ie unfamiliar or unexpected events in a given context~\cite{chandola1,01georg}) in the video which may demand instant action. Due to the rarity and diversity of anomalous events, adequate training anomaly samples are typically not available for supervised training. Hence, researchers in the field dedicated more interest to semi-supervised approaches, in which normals are learned via a proxy task (i.e., a task that indirectly helps to achieve the target goal), and anomalies are detected by finding the deviations from normalities. For example, reconstruction of current frames or prediction of masked frames are popular proxy tasks in video anomaly detection (VAD), in which the trained models on normals show a worse reconstruction or prediction result for anomalies, and the error of the estimation determines the anomaly score. Researchers have employed different proxy tasks in multiple branches to consider different modalities (mostly appearance and motion) in their approaches. Different proxy tasks are meant to be complementary to each other and consequently are combined towards higher performance. For example, Nguyen and Meunier~\cite{02Ngun} proposed a two-stream network in which one stream models the appearance features and detects appearance-based anomalies while the other one models motion patterns and looks for motion anomalies. Multiple similar strategies have been proposed and each work proposes a different combination of proxy tasks with different anomaly score fusion strategies~\cite{01georg,02Ngun,03Memo,04Iones,05Doshi,TANG}. Recently, researchers (\eg \cite{01georg, multi1, folder2}) proposed to add more proxy tasks (\ie multi-task learning based methods) to cover more spatio-temporal patterns. The key question in multi-task learning based methods is how many/what proxy tasks to choose in order to be complementary and to increase performance. Generally, adding more proxy tasks may result in better performance; however, it adds to the computational load and running time. Hence, the design goal is to propose the least number of complementary tasks, considering their abilities and shortcomings in the detection of several anomaly types, in order to cover all necessary attributes. It is worth noting that explainable anomaly detection requires having a strong explanation behind choosing each proxy task. 

\begin{figure*}[h!]
  \centering
   \includegraphics[width=0.60 \linewidth]{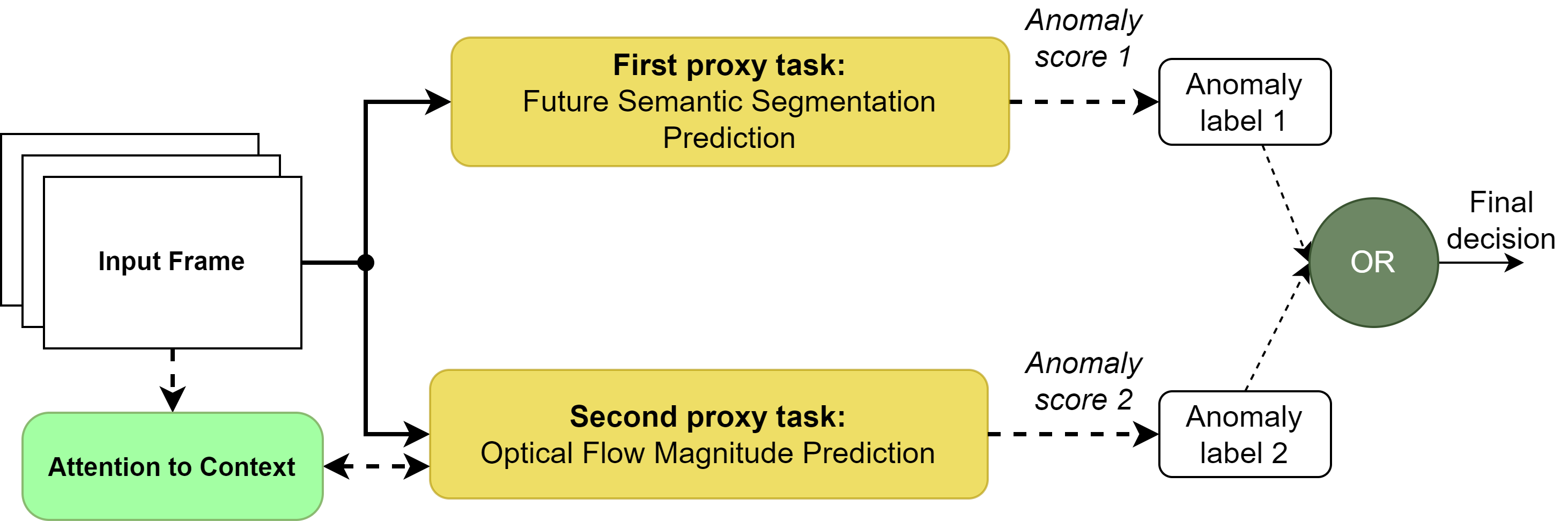}
   \caption{A general view of our proposed method.}
   \label{fig:intropic}
   \vspace{-4mm}
\end{figure*}

Although recent methods attain better results, they still either do not consider motion patterns thoroughly or do not explicitly analyze the class of the object for anomaly detection. To address these shortcomings, inspired by~\cite{01georg}, we propose an improved multi-task learning based VAD method. Different from~\cite{01georg} and following the success of the semantic segmentation task in considering the object class in VAD~\cite{03Memo}, we benefit from the semantic segmentation proxy task's ability, in an improved way in the appearance branch. Additionally, contrarily to~\cite{01georg}, which performs anomaly detection at the object level, we propose a holistic VAD method, avoiding their drawback of losing location information.

Nguyen \etal ~\cite{02Ngun} proposed to estimate optical flow (OF) from a single frame to model motion patterns and to detect related anomalies. However, estimating OF from a single frame could be confusing for the motion network. To overcome this problem, Baradaran and Bergevin ~\cite{03Memo} proposed using optical flow magnitude (OFM) estimation for each object to detect motion anomalies. Although their method addresses the mentioned problem, their method neglects the motion direction information in motion estimation. Hence, their method does not effectively detect the anomalies which are due to sudden direction changes (such as in fighting, jumping, etc). Besides, to make a correspondence between each object and its motion magnitude (i.e., pixel-based object displacement through frames) some important factors, such as motion direction, object part, and distance of the object from the camera were not taken into account.

We propose a new method to address the previous issues. The proposed method leverages two different attention mechanisms. It takes advantage of a spatial and channel attention network and applies it to feature maps of the mid-layers in the encoder, which helps the network consider object parts (hands, feet, etc) for motion magnitude estimation. Moreover, a new attention network is designed that helps estimate the motion magnitude for each object with attention to its distance from the camera and the direction of its motion (details in \cref{sec:method}). Finally, future frame prediction is leveraged, as another proxy task, to find sudden motion changes. In order to reduce the network size, the semantic segmentation and future frame prediction tasks are combined into a novel task of future semantic segmentation prediction to be performed by a single network.

In summary, our contributions are:
\begin{itemize}
\itemsep0em
\item A novel multi-task learning based video anomaly detection method that combines three complementary proxy tasks, "future frame prediction", "semantic segmentation", and "optical flow magnitude prediction", in an explainable way, to more generally consider appearance and motion features for anomaly detection.
\item A combination of semantic segmentation and future frame prediction tasks into a novel proxy task to find both appearance and motion anomalies. To the best of our knowledge, this is the first work that introduces the future semantic segmentation prediction proxy task for video anomaly detection.
\item A novel attention network to estimate precise motion magnitude for an object with attention to its motion direction and its distance from the camera. This introduces a novel way to engage context information in modeling normals and to detect respective anomalies. We also employ a spatial and channel attention mechanism in the backbone of the motion estimation branch to boost meaningful features and generate estimations specific to different object parts.
\end{itemize}

A general view of our proposed method is illustrated in
\cref{fig:intropic}.

\section{Previous work}
\label{sec:litriture}
Researchers have formulated video anomaly detection via various proxy tasks especially frame reconstruction~\cite{06Hasan, 07Chong,08Lerux,09Abati, 10Park, 10Gong, folder4, b40, b41.1, b42, b46, b47, b48, robert3, b51, b52, b53, b54, b56, b50, b55, Luo} or prediction tasks~\cite{11Liu,12Akcay,13Cheng, 14Zhang, 15Zhao, folder5, folder7, folder6, folder8, folder9, folder10, b62, b53.1, b44, b65, b66, traffic, Dong,Liu}, assuming that the unsupervised network (\eg A UNet) trained on normals generates a higher reconstruction/prediction error for anomalies. However, all previously mentioned methods consider low-level features (color, intensity, etc) for anomaly detection and do not explicitly consider the class of objects for their evaluation. Object-centric VAD methods~\cite{04Iones, 17Roy,georgia2, b69, b70, b71, b72,b73} detect and crop objects out of frame (by a pre-trained object detector) but they only consider low-level features in training and inference. Inspired by~\cite{18Lis, 63Bergmann}, Baradaran and Bergevin~\cite{03Memo} proposed a knowledge distillation based VAD that uses semantic segmentation as the proxy task and hence is able to explicitly consider the class of the objects for VAD.

\begin{figure*}[t!]
  \centering
   \includegraphics[width=0.60\linewidth]{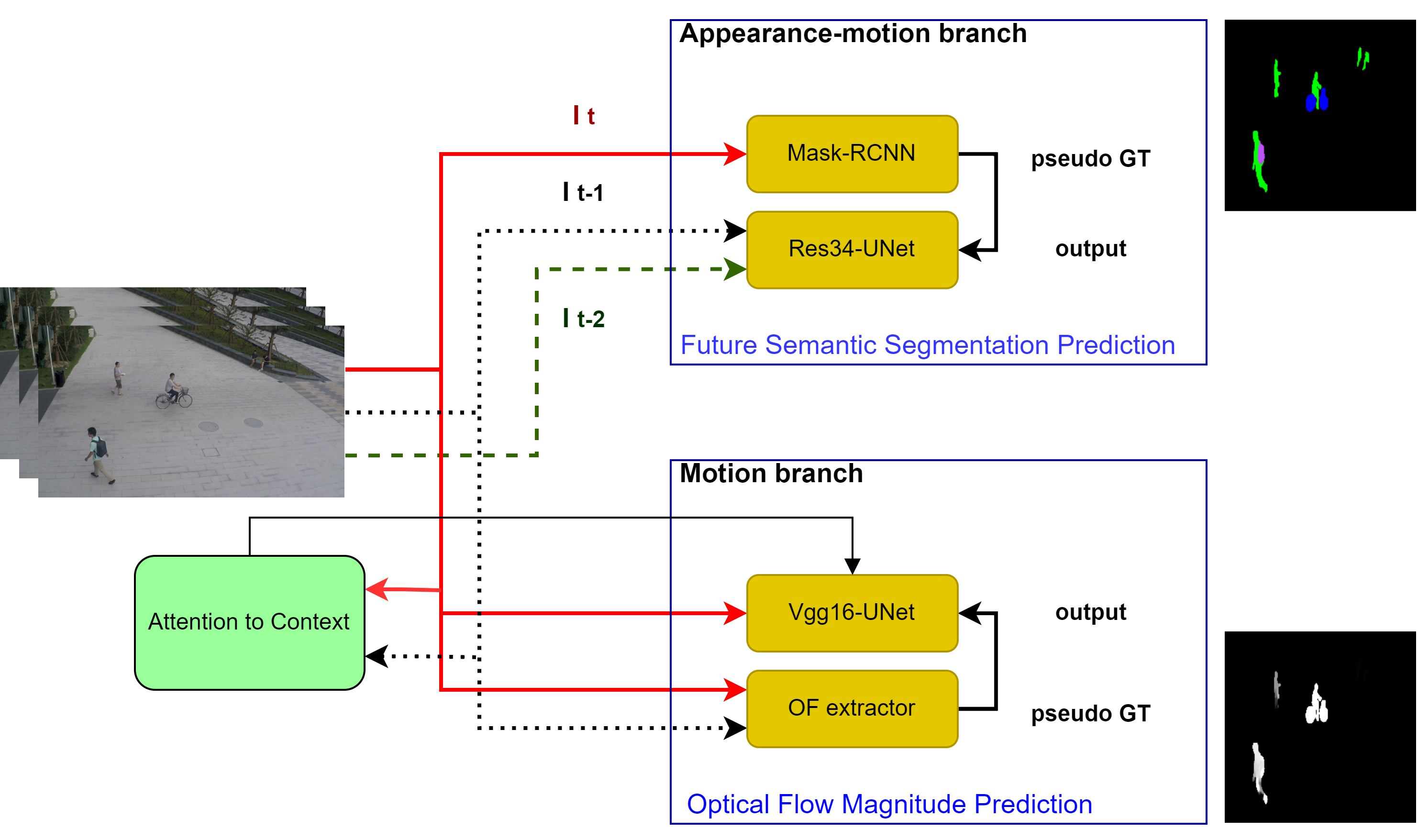}
   \caption{The pipeline of our proposed method. Mask-RCNN and OF extractors (as teachers respectively for the appearance-motion and the motion branches) provide the required pseudo-GT for training their students. During the inference stage, the output of each student network (estimation) is compared to its related pseudo-GT (expectation) to calculate the respective anomaly score. Best viewed in color.}
   \label{fig:short-a}
   \vspace{-3mm}
\end{figure*}

Baradaran and Bergevin~\cite{19Baradaran} report that single-branch approaches (such as~\cite{06Hasan,07Chong}) do not effectively cover all motion cases and are usually dominated by appearance features. Hence, to tackle the shortcomings of one-stream methods, researchers ~\cite{02Ngun,03Memo,04Iones,05Doshi} have proposed two-stream VAD methods, to detect motion and appearance  effectively in separate branches. They mostly tackle the motion anomaly detection problem by reconstructing motion features (\eg optical flow features, two-image gradients, etc.)~\cite{27Liu}. One of the most noticeable related works is proposed by Neygun and Meunier~\cite{02Ngun} which formulates motion learning as a translation from the input frame to its corresponding optical flow map, trying to consider the correspondence between objects and their motion for the motion anomaly detection. Baradaran and Bergevin~\cite{03Memo} proposed to translate the input frame to its optical flow magnitude (considering only the magnitude of motion), as they reported that the network can be confused while predicting the complete optical flow from a single frame. Although their approach addresses the issue of confusion and learns the correspondence between each object and its motion magnitude to detect related anomalies, their method neglects the direction information. Hence their method may fail in the detection of anomalies that are due to sudden direction changes. Moreover, they may fail in precise motion magnitude prediction since the perceived motion in frames is also a function of factors such as the distance from the camera (i.e. the same motion magnitude looks smaller farther) and also the direction of motion (objects moving parallel to the camera look faster than the same motion away from the camera), and these essential factors have not been taken to account in their method.

Inspired by the success of multi-task learning based methods in considering different aspects essential for anomaly detection, we propose an improved multi-task learning based VAD method with complementary proxy tasks to overcome the aforementioned shortcomings and to cover appearance and motion anomalies more effectively. Motivated by the success of future frame prediction and semantic segmentation prediction tasks, respectively in detecting sudden motion changes and object class aware appearance anomalies, we combine them into a single task and introduce the future semantic segmentation prediction as a novel proxy task for video anomaly detection. We also design attentive layers which learn motion considering essential context information and estimate motion with attention to the object part, motion direction, and distance of the object.

\section{Method}
\label{sec:method}
We propose a multi-task learning based video anomaly detection method that leverages three self-supervised proxy tasks abilities in two separate branches in order to model normal patterns and consequently detect anomalies. The pipeline of the proposed method is illustrated in \cref{fig:short-a} and described in detail in the following.

\subsection{Multi-task learning}
Inspired by \cite{01georg} we propose a multi-task learning based VAD method, which leverages  three proxy tasks in two branches for video anomaly detection. The first branch (named appearance-motion branch) combines two different tasks (semantic segmentation and future frame prediction) to model appearance and motion simultaneously. The second branch (\ie the motion branch) is in charge of learning the correspondence between each normal object and its normal motion magnitude, with attention to its distance from the camera, motion direction, and its body parts. In this way, all three tasks are complementary to each other, each trying to find anomalies for which the other tasks may be sub-optimal.

\subsection{The appearance-motion branch}
Experiments in \cite{03Memo} show that leveraging semantic segmentation as a proxy task helps to effectively find the appearance anomalies considering the class of objects. Moreover, our experiments (related information is provided in \cref{sec:exp}) show that future frame prediction is a suitable task to detect sudden motion changes (\eg direction changes, acceleration) since a network trained to predict the future frame of two consecutive normal frames fails to predict the precise location of the objects having sudden motion changes (such as in fighting, jumping, etc). To leverage the abilities of both tasks, the first branch of our method combines two different tasks of semantic segmentation and future frame prediction in a new single task and aims to predict the semantic segmentation map of the future frame by observing two consecutive frames. In this way, not only does it learn the class of the normal objects in the frame during the training, but it also learns the normal evolution between two normal frames. This branch follows a teacher-student strategy for anomaly detection. During training, a student (resnet34-UNet in our method) gets two consecutive frames and learns to generate the semantic segmentation map of the future frame (i.e., the next frame), assuming that at inference time, the prediction error would be higher for anomalies. The pseudo Ground-Truth (GT) for training the student network is generated by Mask-RCNN which is trained on MS COCO.

\subsection{The motion branch}

Although the first branch partially considers motion anomalies, it does not cover all motion cases. Hence, in the second branch, we employ the idea proposed in ~\cite{03Memo} which is to learn a normal motion magnitude for each object by translating an input frame to its optical flow magnitude map. This branch also follows a teacher-student strategy where the student (a vgg16-UNet here) learns to translate an input frame to its optical flow magnitude map, generated by a pre-trained optical flow extractor (considering its past frame) as a pseudo-GT. In this way, the student network learns the correspondence between each object and its normal motion magnitude during training, assuming that it will make an imprecise motion estimation for objects moving faster/slower than their normal motions. However, the original method has some challenges as follows: 1) some objects (such as humans) do not have a constant motion magnitude in all their body parts. For example, hands and feet usually have a larger motion magnitude compared to the chest and head. This factor has not been taken into account in ~\cite{03Memo}. 2) The motion magnitude perceived in frames (pixel displacement of objects) is a function of some variables such as motion direction and object distance from the camera. Objects moving parallel to the camera show a larger motion (i.e, generates a larger optical flow magnitude) compared to objects moving away/towards from/to the camera. Hence motion modeling and estimation of motion magnitude without attention to these factors would result in an imprecise prediction.

\begin{figure}[ht]
  \centering
   \includegraphics[width=1\linewidth]{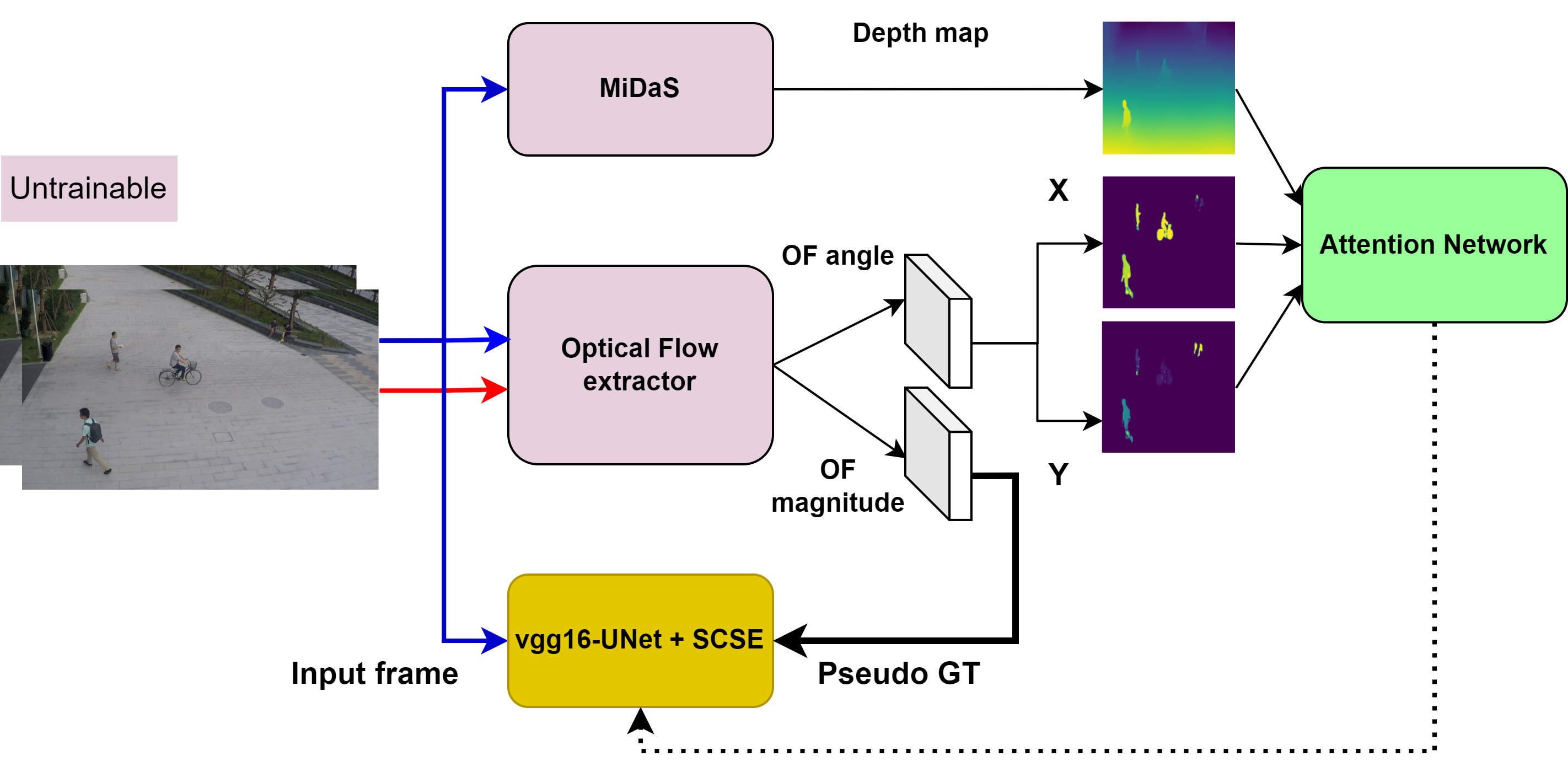}
   \caption{The motion branch in detail. Best viewed in color.}
   \label{fig:short-b}
   \vspace{-4mm}
\end{figure}

In our motion branch (\cref{fig:short-b}), we employ two different attention mechanisms to address the mentioned shortcomings and engage essential factors in motion modeling to address the shortcomings in ~\cite{03Memo}. We employ a spatial and channel attention network in the main network, to dedicate more attention to special body parts (such as feet and hands) and we also design a new attention network to help the network make predictions with attention to supplementary information (such as motion direction and relative distance information). The details of the attention mechanisms are provided next.

\subsection{Spatial and channel attention}
By visualizing the generated feature maps of the basic vgg16-UNet in the motion branch, we observed that the encoder of the vgg16-UNet (trained semi-supervisedly) generates different levels of features through different layers and the mid-layers generate feature maps that are activated for different object parts. Hence, we applied the spatial and channel attention (SCSE) mechanism proposed in ~\cite{26Groy} on the feature maps of mid-layers to help the network dedicate more attention to different body parts. 

\subsection{Attention to distance and direction}
\begin{figure}[ht]
  \centering
   \includegraphics[width=1.0\linewidth]{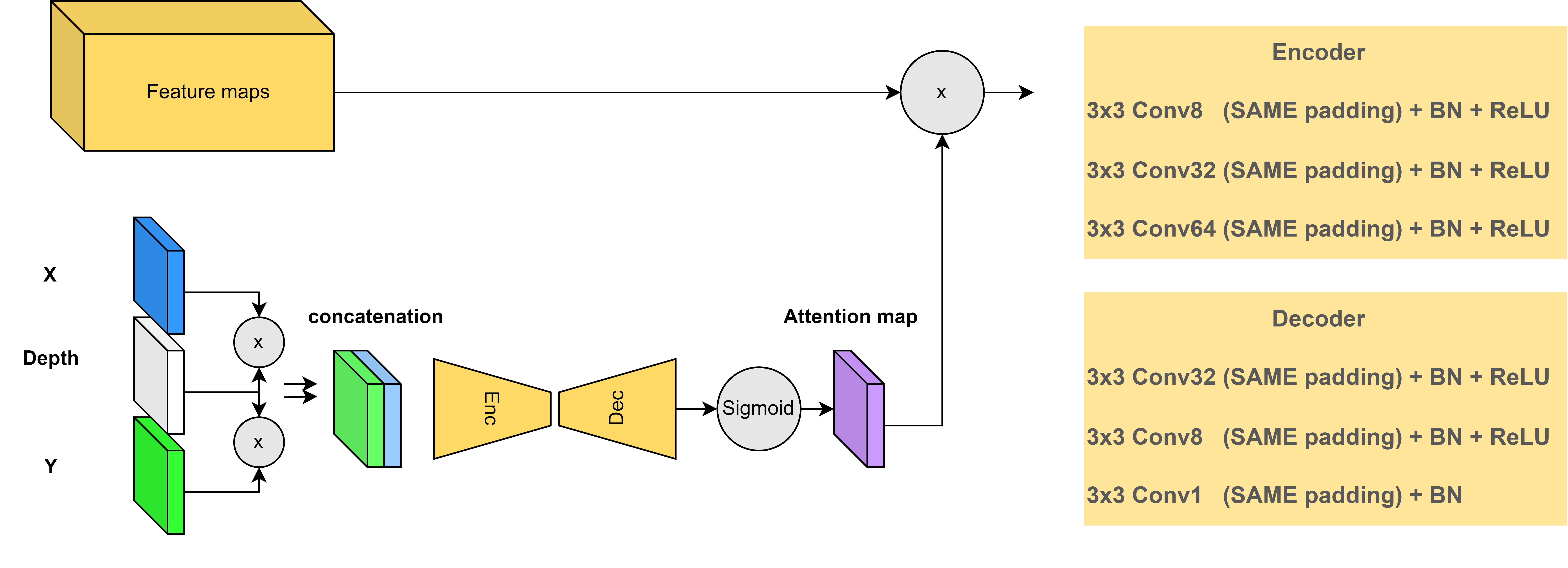}
   \caption{The proposed attention network. The generated attention map is applied to feature maps to provide attention to supplementary information (i.e., depth and direction). Best viewed in color.}
   \label{fig:attention}
   %\vspace{-4mm}
\end{figure}

Normality is defined in context. Hence the essential point and challenge in semi-supervised anomaly detection methods is to model normal patterns precisely and to find anomalies by measuring deviations from normals. Previous methods do not consider important factors (such as direction and distance from camera) for motion modeling (and estimation). To provide attention to these factors and obtain a precise motion model for objects, we designed another attention network, as illustrated in \cref{fig:attention}. The network uses the direction and distance information as inputs and generates an attention map to apply to attentive layers.

In the motion branch, the teacher network extracts the optical flow of two consecutive frames (${I_{t-1}}$, ${I_{t}}$) and provides the magnitude of the optical flow as the pseudo-GT to be utilized by the student to map the input frame (${I_{t}}$) to its OFM. The generated OF features by the teacher (denoted as OF in \cref{eq:eq1}) encompasses the direction information in addition to magnitude information. In \cref{eq:eq1} \verb'Mag' stands for the magnitude of the optical flow and \verb'Ang' stands for the angle of the motion relative to the horizontal axis in the frame. The \verb'Ang' features can be supplied to our attention network as direction features. However, we calculate the Cosine and Sine of \verb'Ang' to normalize it and to generate two different features: motion parallel to the camera (denoted as X in \cref{eq:eq2} and \cref{fig:attention}) and motion towards/away to/from the camera (denoted as Y in \cref{eq:eq3} and \cref{fig:attention}).

\begin{equation}
  Mag, Ang = OF(I_{t-1},I_{t})
  \label{eq:eq1}
  \vspace{-3.5mm}
\end{equation}

\begin{equation}
  X = |Cos(Ang)|
  \label{eq:eq2}
  \vspace{-3.5mm}
\end{equation}

\begin{equation}
  Y = |Sin(Ang)|
  \label{eq:eq3}
\end{equation}

%In (2,3), X and Y includes the direction information related to two orthogonal axises (i.e, X and Y). The abs function calculates the absolute of the Sine of Cosine values respectively in (2) and (3).
Since we do not have the actual information about the object's distance from the camera, we extract the depth maps of the frames to represent the relative distance information of the objects to the camera. We use MiDaS ~\cite{21Ren1, 22Ren2} to estimate the relative depth maps of input frames. MiDaS ensures high-quality depth map generation for a wide range of inputs since it is pre-trained on 10 different datasets using multi-objective optimization. We use the hybrid version of the method to balance precision and execution time. \Cref{fig:attention} shows how the extracted informations are combined and processed inside the attention network to generate the attention map.

\subsection{Inference}
At the inference stage, we provide both normal and abnormal frames to each branch and we compare the estimation of each branch with their respective expectations (\ie pseudo-GT generated by teachers of each branch) to calculate the anomaly map of that frame. For each of the two branches, we calculate the sum of activations in the anomaly map as the anomaly score S(t) of that branch for the frame (\cref{eq:eq4}).

\begin{equation}
  S(t)= \Sigma |Out_{student}(I_{t})-Out_{teacher}(I_{t})|
  \label{eq:eq4}
\end{equation}

In \cref{eq:eq4}, $Out_{student}(I_{t})$ and  $Out_{teacher}(I_{t})$ respectively denote the estimations and expectations of a given branch. Summation is done over all pixels in the anomaly map (\ie the difference between student and teacher outputs).

In our experiments, we found some false positives which were due to two reasons: 1) false detections or misdetections by Mask-RCNN which produce large activations in the anomaly map. 2) jumps between frames which are apparently due to recording or saving issues. These jumps generate false large motions between some frames. The mentioned false positives produce sudden jumps/falls in the anomaly score of some frames. However, considering the frame rate of the video, we assume that adjacent frames should have a similar anomaly score. Hence, to relax the anomaly scores (\ie temporal denoising), Savitzky–Golay filter (\cref{eq:eq33})~\cite{23Chongke} is applied on the anomaly scores. 

\begin{equation}
  S_{r}(t)= \frac{1}{N} \mathlarger{\mathlarger{\sum}}_{i=-w}^{i=w} \alpha^{S(t+i)}
  \label{eq:eq33}
 \end{equation}

In this equation, $S_{r}(t)$ represents the relaxed anomaly score generated from noisy anomaly scores $S(t)$. N is the normalizing factor and $\alpha$ and $w$ are the convolutional coefficients and window size respectively.

Finally, as the final decision, we flag a frame as an anomaly if and only if the anomaly score S(t) of either branch 1 or branch 2 (or both) is larger than a predefined threshold. As the networks of each branch have been trained with normal frames, we expect to observe a considerable difference between estimations and expectations if the input frame contains any anomaly specific to that branch. These anomaly maps are expected to contain activations at the position of the anomalies in the frame.

\section{Experiments and results}
\label{sec:exp}
We trained and evaluated the performance of our proposed method and the effectiveness of each contribution on the ShangahiTech Campus ~\cite{11Liu} and UCSD-Ped2 ~\cite{24Mahadevan} datasets. The details of the experiments,  qualitative and quantitative results, and a comparison with state-of-the-art approaches appear next.

\subsection{Datasets}
ShanghaiTech Campus and UCSD-Ped2 datasets are two of the benchmark datasets popularly used to evaluate semi-supervised VAD methods. They provide only normal frames in their training subsets and both normal and abnormal frames in their test subsets, along with frame-based and pixel-based annotations. The definition of normality and anomaly is similar in both datasets. People walking on the sidewalk (could be carrying bags or backpacks) are considered normal, however, the presence of some previously unseen objects (such as bikes, bicycles, cars, etc) or some previously unseen motion patterns (such as running, chasing, fighting, riding, etc) are considered as anomalies. Compared to UCSD-Ped2, ShanghaiTech Campus is a more complex dataset, since it has multiple different scenes (13 scenes) and a larger number of anomalies. On the other hand, low resolution and gray scale frames make the UCSD-Ped2 dataset challenging and prone to failures for the segmentation task.

\subsection{Evaluation metric}
Following the state-of-the-art (SOTA) methods in the field, we provide our quantitative evaluation by measuring the frame-level AUC (Area Under Curve). This curve is plotted by registering multiple True Positive Rate (TPR) and False Positive Rate (FPR) of the method by changing the anomaly score threshold from min to max. A higher AUC indicates better performance.

\subsection{Implementation details}
Input frames in our experiments are resized to 256*256 for each branch. For a faster convergence, we initialize encoders of both student networks (res34-UNet and vgg16-Unet) with parameters of the networks trained on Imagenet. The learning rates of both branches are initialized to 0.001 and are halved every 10 epochs. We trained networks of both branches with the patch-based MSE loss (\cref{eq:patch})~\cite{03Memo} (dividing the input frame into 16 patches) by the Adam optimizer. We employed Mask-RCNN (pre-trained on MS COCO) as the teacher of the appearance branch and the Farneback algorithm from the OpenCV library as the teacher for pseudo-GT optical flow feature extraction from two consecutive frames. Finally, to discard the background and to bring more attention to the foreground objects, we mask each extracted optical flow map with the corresponding semantic segmentation map.

\begin{equation}
\text{Patch-loss} = \max (Loss_{i}) 
\label{eq:patch}
\vspace{-4mm}
\end{equation}
where:

$Loss_{i}$  is the MSE loss in $i_{th}$ patch in the frame.

\subsection{Future frame prediction}
\begin{figure}[t]
  \centering
   \includegraphics[width=0.72\linewidth]{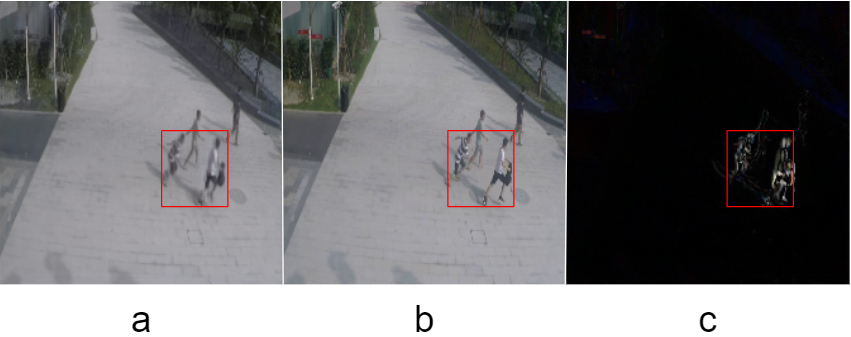}
   \caption{(a): Predicted future frame. (b): Actual future frame. (c): Difference between the predicted and actual future frame.}
   \label{fig:prediction_BB}
   \vspace{-4mm}
\end{figure}

\begin{figure}[t]
  \centering
   \includegraphics[width=0.42\linewidth]{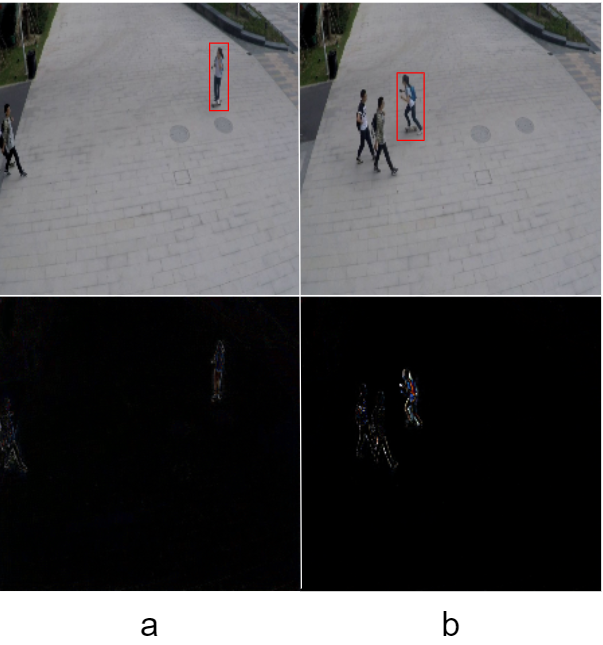}
   \caption{Future frame prediction task’s ability in detecting fast motions vs detecting sudden direction changes. (a) Future frame prediction fails in the detection of abnormal motion with constant speed. (b) Future frame prediction task detects abnormal motions with sudden motion changes. The brightness of the anomaly maps has been increased slightly to better show the activations.}
   \label{fig:anomaly_gen}
   \vspace{-4mm}
\end{figure}

To explore the abilities of future frame prediction proxy tasks in more detail, we conducted a preliminary experiment. We trained resnet-UNet to estimate future frames by observing two consecutive frames. The qualitative results (\Cref{fig:prediction_BB}) demonstrate the ability of the future frame prediction proxy task in finding the sudden motion changes. As seen in this figure, the estimated and the real future frames are different in the position of chasing persons (hence producing larger activations in the anomaly map). However, the network is able to generate a precise prediction, at the position of objects with normal motion.
It is worth noting that, as this proxy task is trained on normal speed motions it can find abnormal fast motions as well. However, due to the high capacity of CNNs, the future prediction ability can be generalized to these anomalies too.
In other words, the CNN can predict the precise position of fast objects if they move monotonically fast through frames and don’t show sudden changes in direction. In \Cref{fig:anomaly_gen}(a), a skate rider travels fast but uniformly through adjacent frames, hence the CNN does not produce a high anomaly
activation in that position. However, as she suddenly puts her foot on the ground to accelerate (\Cref{fig:anomaly_gen}(b)), the generated anomaly activation is comparatively higher. In our proposed method, this shortcoming is handled by the OFM prediction proxy task.

\subsection{Qualitative evaluation}
\begin{figure}[t]
  \centering
   \includegraphics[width=0.9\linewidth]{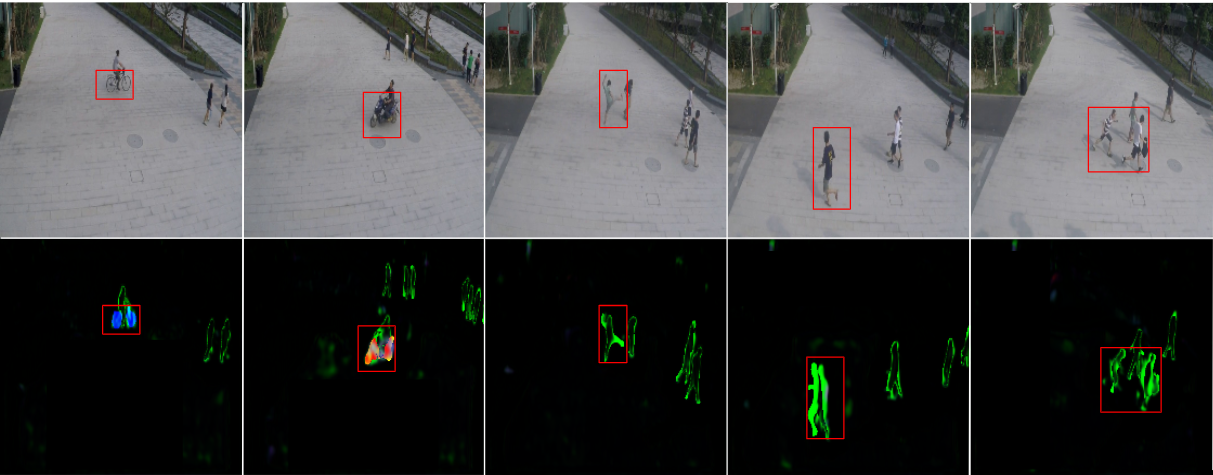}
   \caption{Qualitative results for the appearance-motion branch. Top: Input frames. Bottom: Anomaly maps. Anomalies are indicated by red bounding boxes.}
   \label{fig:app-mo}
   \vspace{-3mm}
\end{figure}

\begin{figure}[t]
  \centering
   \includegraphics[width=0.8\linewidth]{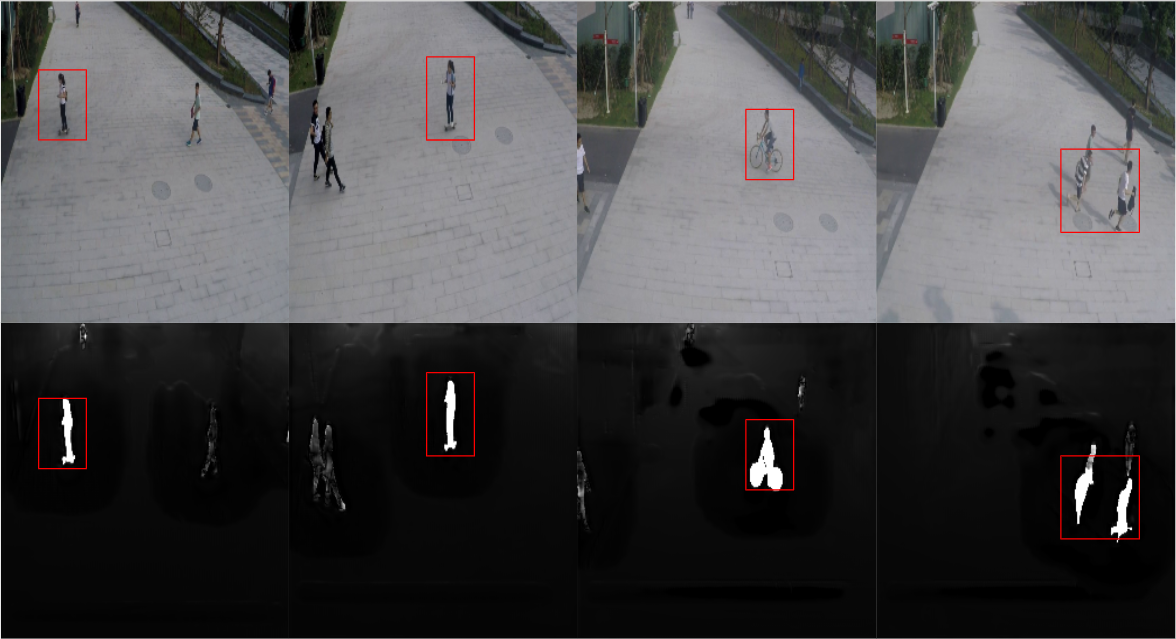}
   \caption{Qualitative results for the motion branch. Top: Input frames. Bottom: Anomaly maps. Anomalies are indicated by red bounding boxes.}
   \label{fig:motion}
   \vspace{-3mm}
\end{figure}

To qualitatively analyze the effectiveness of the proposed method, we observe the anomaly maps for multiple normal and abnormal frames. \Cref{fig:app-mo,fig:motion} present qualitative results for the appearance-motion and motion branches, respectively. \Cref{fig:app-mo} contains multiple samples of anomalous frames (top row) and corresponding anomaly maps (bottom row), generated by the appearance-motion branch. As can be seen in the figure, anomaly maps contain higher activations at anomalous objects (bicycle and motorcycle in columns 1 and 2) or even at normal objects with sudden abnormal motions (legs and hands of a fighting man in column 3 and for running or chasing persons in columns 4 and 5).

 Similar results can be observed in \cref{fig:motion} for the motion branch. As can be seen, our method generates larger activations for objects with anomalous motions since the estimation of the motion branch is considerably different from its expectation for an anomaly.
 
\begin{figure}[t]
  \centering
   \includegraphics[width=0.8\linewidth]{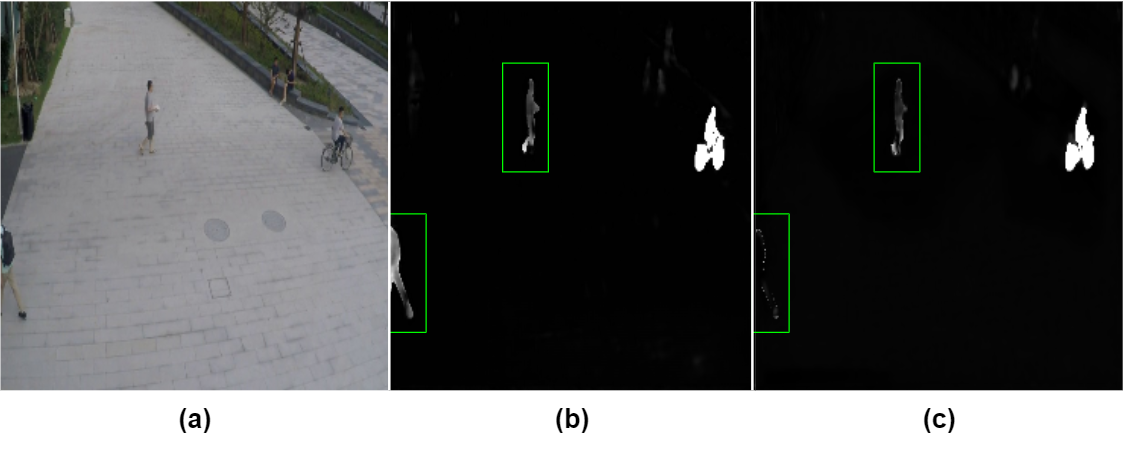}
   \caption{Attention in motion estimation. (a) Input frame. (b) Anomaly map without attention. (c) Anomaly map with attention. Normal moving objects are indicated by green bounding boxes.}
   \label{fig:att-mo}
   \vspace{-3mm}
\end{figure}

\subsubsection{Importance of attention}

\Cref{fig:att-mo} presents an anomalous frame (9a) and the generated anomaly map with and without applying the attention mechanisms. As can be noticed, the generated anomaly map contains weaker activations at the position of normal moving objects (green bounding boxes) in presence of attention compared to without attention. It demonstrates that the motions of normal objects are more precisely estimated when the attention mechanisms are active. For example, for the pedestrian close to the camera (the lower green bounding box in Figure 9b) the basic network does not take the distance information into account and estimates a smaller motion compared to its pseudo-GT which leads to a bigger difference between estimation and pseudo-GT (and hence a larger activation in the anomaly map compared to when the attention mechanisms are active (Figure 9c)). Additionally, we observe a larger activation at the feet of the normal moving object in the upper green bounding box when the attention module is not active. In both cases, the attention mechanisms reduce the likelihood of false anomaly detection. 

\begin{figure}[hbt!]
  \centering
   \includegraphics[width=0.9\linewidth]{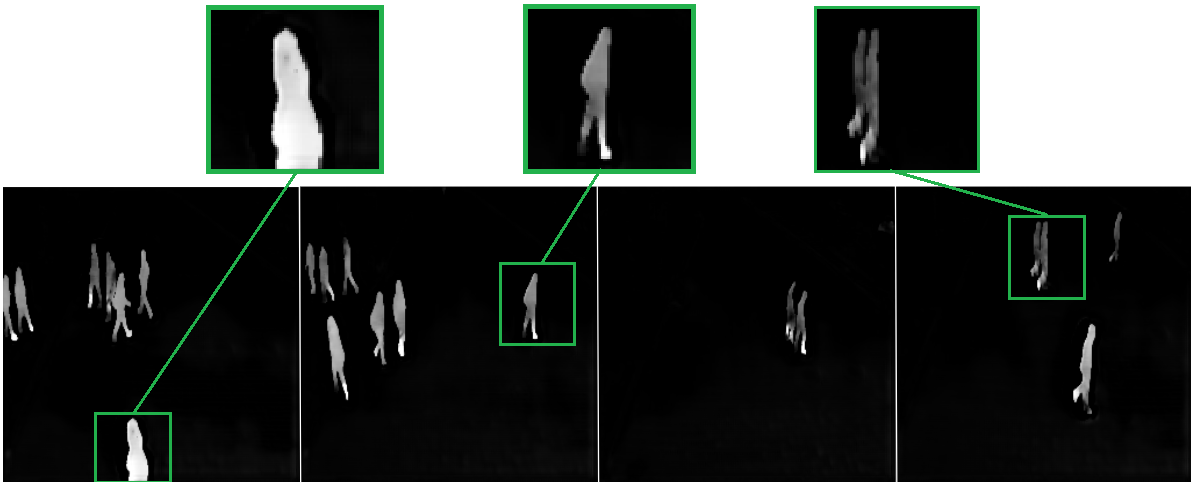}
   \caption{Estimated outputs of the motion branch with attention.}
   \label{fig:after}
   \vspace{-2mm}
\end{figure}

\Cref{fig:after} shows the estimations of the motion branch with attention. As can be observed, the network is aware of the object parts and estimates a larger motion for the feet, compared to other parts. Moreover, the estimated motion is larger for objects close to the camera.

\subsection{Quantitative evaluation}
Table 1 compares the performance of our proposed method with state-of-the-art (SOTA) holistic semi-supervised VAD methods on two benchmark datasets (ShanghaiTech Campus and UCSD-Ped2). The comparison is based on frame-level AUC. To have a fair comparison, we kept the settings of the original papers.

As can be seen, our method provides superior results compared to SOTA holistic semi-supervised VAD methods. Noticeably, the performance improvement (compared to previous methods (especially ~\cite{03Memo})) is more significant for the ShanghaiTech Campus compared to the UCSD-Ped2. This can be explained by the fact that precise motion modeling is more crucial in ShanghaiTech Campus as it has objects at different distances from the camera and motions in various directions, compared to UCSD-Ped2 which is limited in terms of motion directions and distances from the camera. Hence, our contributions in motion modeling (i.e. engaging context information by attention mechanisms) bring larger improvements for ShanghaiTech Campus, confirming our method successfully addresses the identified limitations of previous methods.

\begin{table}[h!]
\begin{center}
\scalebox{0.95}{
\begin{tabular}{l|c|c}
\hline
Method & ShanghaiTech & UCSD \\% // & Avenue \\
  & Campus & Ped2\\
\hline\hline
Hasan \etal \cite{06Hasan} & 60.9 & 90.0 \\%  & 70.2 \\
Chong \etal \cite{07Chong} & N/A & 87.4 \\%  & 80.3\\
Dong \etal \cite{Dong} & 73.7 & 95.6 \\%  & 84.9\\
Liu \etal \cite{11Liu} & 72.8 & 95.4 \\%  & 85.1\\
Liu \etal \cite{Liu} & N/A & 87.5 \\%  & 84.4\\
Park \etal \cite{10Park} & 70.5 & 97.0 \\%  & 88.5\\
Gong \etal \cite{10Gong} & 71.2 & 94.1 \\%  & 83.3\\
Nguyen \etal \cite{02Ngun} & N/A & 96.2 \\%  & 86.9\\
%Ionescu \etal \cite{04Iones} & 84.9 & \textbf{97.8} & 90.4\\
Tang \etal \cite{TANG} & 73.0 & 96.3  \\% & 85.1\\
Baradaran \etal \cite{03Memo} & 86.18 & 97.76  \\% & N/A\\
Georgescu \etal \cite{01georg} & 83.5 & 92.4 \\%  & 86.9\\
Luo \etal \cite{Luo} & 73.0 & 95.0  \\% & 86.8\\
Yu \etal \cite{yu1} & 73.0 & 95.0 \\%  & 86.8\\
\hline
Ours & \textbf{89.1} & \textbf{97.8} \\%  & --\\
\hline
\end{tabular}}
\end{center}
\caption{Performance comparison (frame-level AUC) with SOTA methods. The best-performing method is denoted in boldface.}
\vspace{-4mm}
\end{table}

The above qualitative and quantitative results are sufficient to show our goal is attained. Nevertheless, it could be interesting to further assess the generality of our method with experiments on other datasets and one possibility we investigated is the Avenue dataset, which is commonly used in the field. Although comparable with SOTA methods, the results obtained are not as meaningful as those in Table 1, for two main reasons: 1) concept of anomalies (based on the available annotations) in the Avenue dataset is not totally compatible with the goal and formulation of our method, resulting in a number of invalid true positives and false positives. 2) low camera height results in the generation of  considerable scene occlusions with the ensuing difficulty in properly estimating the motion direction, which is needed to address the limitations of previous methods.

\subsection{Ablation study}

Tables 2, 3, and 4 confirm the effectiveness of each proposed contribution in increasing the obtained performance. As can be noticed in Table 2, all proxy tasks are complementary to each other and can detect more anomalies when combined. Most importantly, we observe that by adding the future frame prediction task to our method, it detects more motion anomalies compared to just using the OFM.

Table 3 demonstrates quantitatively the contribution of attention mechanisms in the proposed method. It is worth noting that in order to concentrate on the contribution of each attention mechanism in improving the performance of motion modeling and motion anomaly detection, we have conducted this ablation study only on the motion branch. As shown in the results, adding attention mechanisms to the motion branch results in higher performance.

\begin{table}[hbt]
\begin{center}
\scalebox{0.83}{
\begin{tabular}{|l|c|c|c|c|c|}
\hline
Tasks & Seg & OFM & Seg+Pred & Seg+OFM & Seg+OFM+Pred \\
\hline\hline
AUC & 76.3 & 79.19 & 80.61 & 88.21 & \textbf{89.1} \\
\hline
\end{tabular}}
\end{center}
\caption{Contribution of each proxy task (Seg: semantic segmentation, OFM: optical flow magnitude, Pred: prediction) in increasing the performance.}
\vspace{-3mm}
\end{table}

\begin{table}[hbt]
\begin{center}
\scalebox{0.8}{
\begin{tabular}{|l|c|c|c|c|}
\hline
Model & UNet & UNet+Att & UNet+Att+SCSE \\
\hline\hline
AUC & 69.7 & 78.14 & \textbf{79.19} \\
\hline
\end{tabular}}
\end{center}
\caption{Contribution of each attention mechanism (Att: attention to supplementary information) in increasing the performance of the motion branch, SCSE: spatial and channel-wise attention \cite{26Groy}. In this ablation study only the motion branch is active (the appearance-motion branch is deactivated).}
\vspace{-3mm}
\end{table}

\begin{table}[ht]
\begin{center}
\scalebox{0.75}{
\begin{tabular}{|l|c|c|c|c|c|}
\hline
Position & No map & Encoder & Decoder & Skip connection & Final layer \\
\hline\hline
AUC & 69.7 & 77.91 & \textbf{78.14} & 72.07 & 77.52 \\
\hline
\end{tabular}}
\end{center}
\caption{Position of attention map. This table shows how the position of attention map integration affects the performance of the motion branch. In this ablation study, only the motion branch with the context attention is active. (The appearance-motion branch and SCSE mechanism are deactivated).}
\vspace{-4mm}
\end{table}

Finally, we conducted another ablation study (Table 4) to analyze the effect of the position of attention maps on the performance of the motion branch. To concentrate on the importance of this factor, we deactivated the appearance-motion branch and the SCSE mechanism in this study. This table indicates that adding attention at any position in the network (encoder, decoder, skip connection, or the final layer) improves performance. However, a higher performance has been observed for the encoder, decoder, and final layer positions, compared to the skip connection.

\section{Conclusion}
\label{sec:conclusion}
We proposed an improved multi-task learning based video anomaly detection method, which introduces future semantic segmentation prediction as a novel proxy task for video anomaly detection and combines multiple complementary proxy tasks for a better consideration  of appearance and motion anomalies. Moreover, we introduced a novel mechanism to improve the precision of motion modeling with attention to context. Experimental results show that adding each proxy task results in higher performance in terms of AUC. Importantly, experimental results confirm that our proposed idea of paying attention to both direction of object motion and the distance of the object from the camera introduces a new and meaningful way to engage context information in video anomaly detection and results in more precise motion estimation and likely fewer false detections. Our qualitative results show the explainability of estimations and detections and also the effectiveness of each contribution. Quantitative results on the ShanghaiTech Campus and UCSD-Ped2 datasets demonstrate the superior performance of our method, compared to SOTA methods.

%%%%%%%%% REFERENCES
{\small
\bibliographystyle{ieee_fullname}
\bibliography{CVPRW}
}

\end{document}